\documentclass{bmvc2k}

\title{Unbiased Image Style Transfer}

\addauthor{Hyun-Chul Choi}{pogary@ynu.ac.kr}{1}
\addauthor{Minseong Kim}{tyui592@ynu.ac.kr}{1}

\addinstitution{
 ICVSLab.\\
 Yeungnam University\\
 Kyungsan, Republic of Korea
}

\runninghead{Prof, Student}{Arxiv e-print}

\def\etal{\emph{et al}\bmvaOneDot}

\begin{document}

\maketitle

\begin{abstract}
Recent fast image style transferring methods use feed-forward neural networks to generate an output image of desired style strength from the input pair of a content and a target style image.
In the existing methods, the image of intermediate style between the content and the target style is obtained by decoding a linearly interpolated feature in encoded feature space. However, there has been no work on analyzing the effectiveness of this kind of style strength interpolation so far.
In this paper, we tackle the missing work on the in-depth analysis of style interpolation and propose a method that is more effective in controlling style strength. We interpret the training task of a style transfer network as a regression learning between the control parameter $\alpha \in [0.0, 1.0]$ and output style strength. In this understanding, the existing methods are biased due to the fact that training is performed with one-sided data of full style strength ($\alpha = 1.0$). Thus, this biased learning does not guarantee the generation of a desired intermediate style corresponding to the style control parameter of $0.0 \leq \alpha < 1.0$.
To solve this problem of the biased network, we propose an unbiased learning technique which uses unbiased training data and corresponding unbiased loss for $\alpha = 0.0$ to make the feed-forward networks to generate a zero-style image, i.e., content image when $\alpha=0.0$. Moreover, with additional anchor data and loss for $0.0 < \alpha < 1.0$, our method allows the network to learn the desired regression consistent with a specific interpolation function in encoded feature space.
Our experimental results verified that our unbiased learning method achieved the reconstruction of a content image with zero style strength, better regression specification between style control parameter and output style, and more stable style transfer that is insensitive to the weight of style loss without additive complexity in image generating process.
\end{abstract}

\section{Introduction}
\label{sec:introduction}
\begin{figure}
\includegraphics[trim={0 15cm 0 0},width=1.0\textwidth]{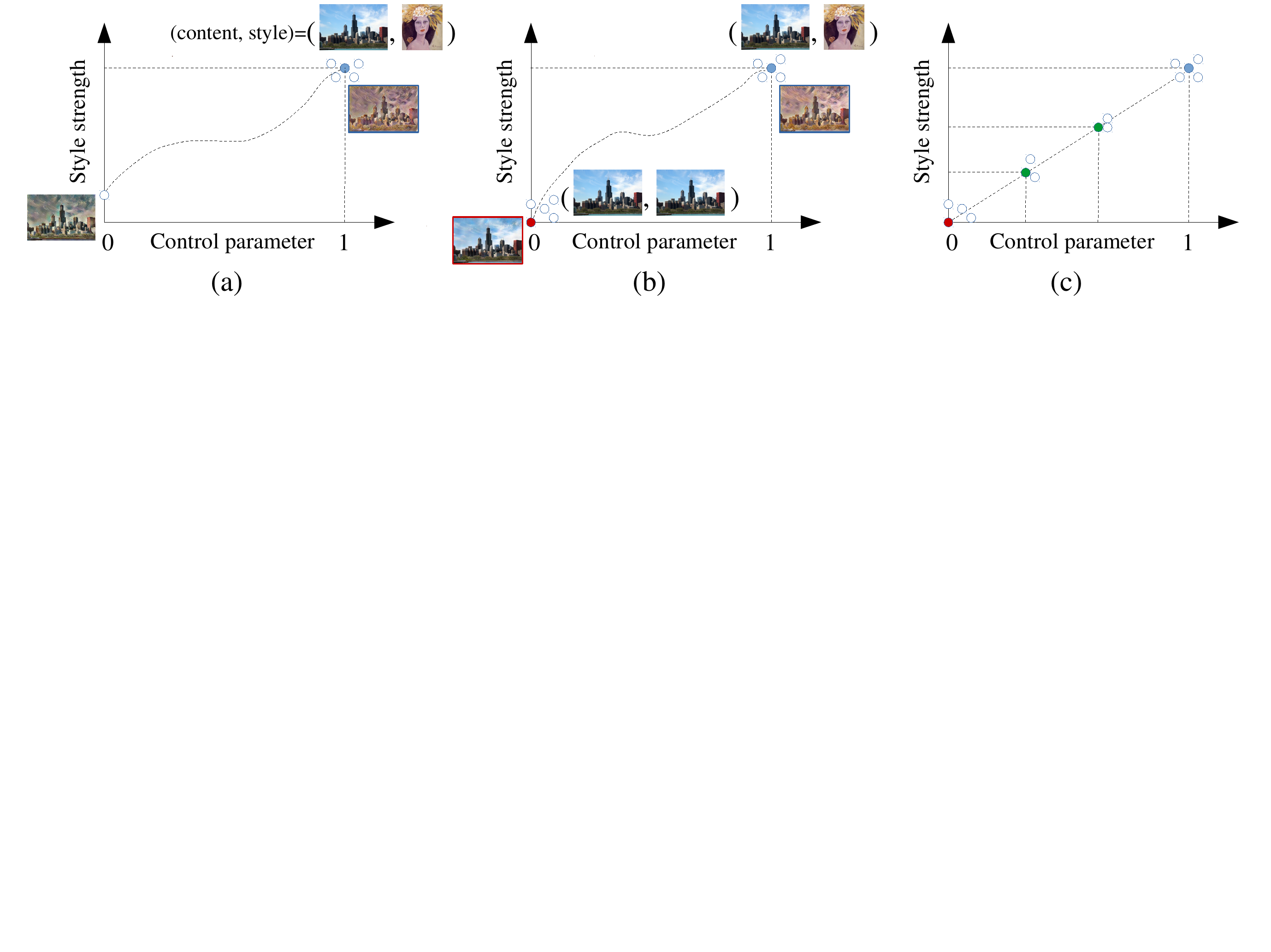}
\caption{Overall concept of our unbiased image style transfer: (a) Previous methods learned regression between control parameter and output style strength by training style transfer networks only with full style data (blue dot).; (b) Our unbiased method uses unbiased data (red dot) to prevent the trained networks from overfitting into the full style data.; (c) Moreover, our method allows the style transfer networks to learn a specific regression function with additional training data (green dots) of intermediate style strengths.}
\label{fig:concept}
\end{figure}

Recent fast image style transferring methods~\cite{Dumoulin_2017_ICLR,Ghiasi_2017_BMVC,Li_2017_IJCAI,Huang_2017_ICCV,Li_2017_NIPS} uses feed-forward networks to generate output stylized image from an input content image or the input pair of a content image and a target style image. Here, the feed-forward networks were trained to learn how to encode feature to represent content and style of an image (encoder), how to change the style of image in feature space (transformer), and how to generate an image from the style-changed feature (decoder). Those approaches also utilize linear interpolation technique to generate images of intermediate style between content image and target style image corresponding to a style control parameter $\alpha \in [0.0, 1.0]$. Although they achieved good results in both processing speed and style quality of output stylized image, it is not guaranteed that the output intermediate style from the linear interpolation in encoded feature space is correctly matched to the desired style strength because the networks were trained only with the biased pairs of a content image and a target style image of full style strength ($\alpha = 1.0$) as shown in fig.\ref{fig:concept}(a). To date, such problem was not dealt in depth.

\begin{figure}
\includegraphics[trim={0 16cm 0 0},width=1.0\textwidth]{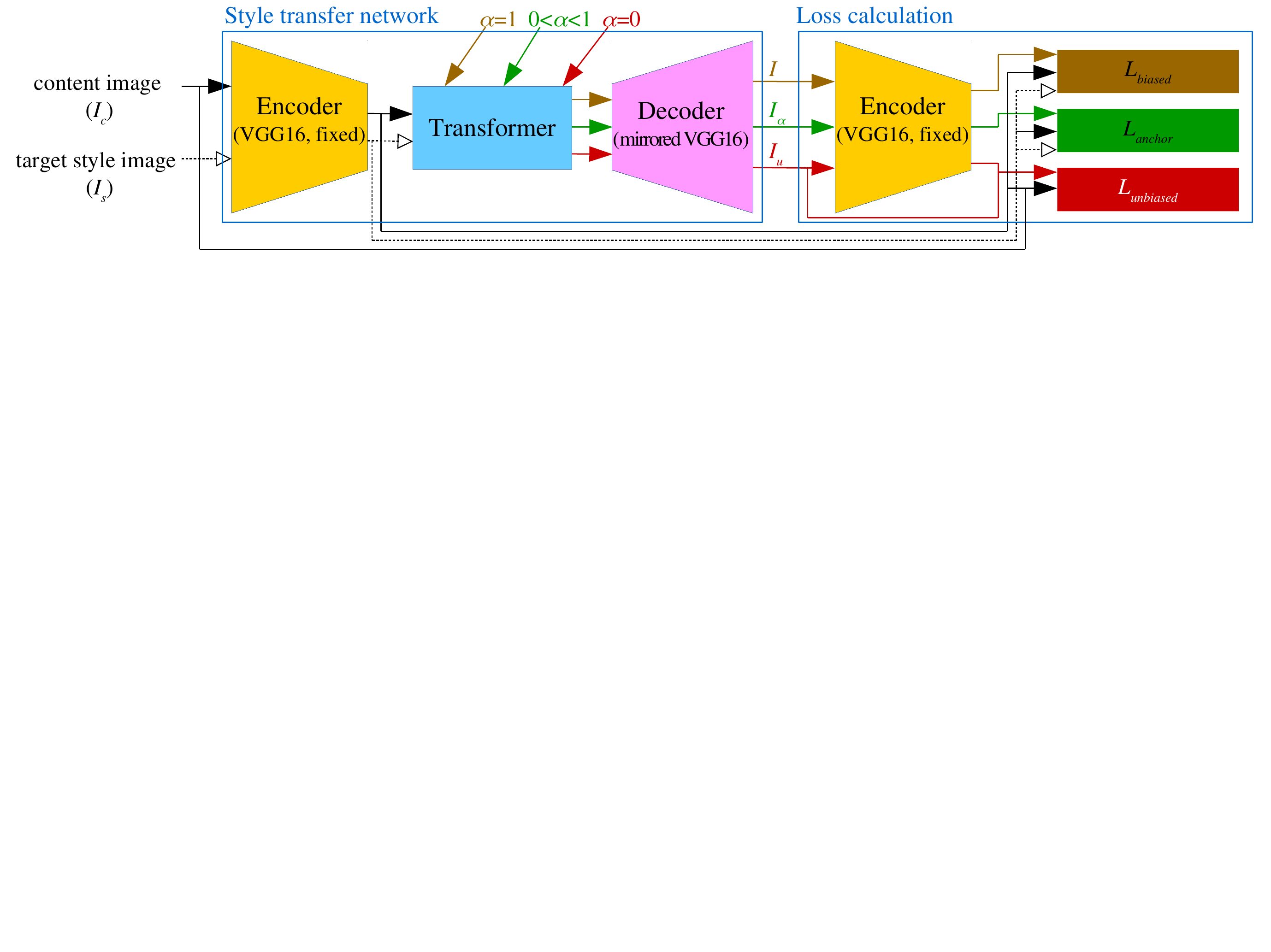}
\caption{Network architecture and losses of our unbiased image style transfer: Our style transfer network has encoder-transformer-decoder architecture. The encoder and the decoder have the symmetric structures based on VGG16 feature extractor. The transformer aligns the content feature by using target style feature like CIN or AdaIN layer, and also controls the output style strength by using a specific interpolation technique with a scalar value $0.0 \leq \alpha \leq 1.0$. For the unbiased learning, in addition to the previously used total loss $L$ (the summation of content loss and style loss), we use $L_{unbiased}$ to reconstruct content image when $\alpha=0.0$ and $L_{anchor}$ to match the learned regression to a specific function.}
\label{fig:network architecture}
\end{figure}

In this paper, we tackle the problem of style interpolation caused by the biased network training and propose a method that is superior to the current linear interpolation technique. First, we interpret the task of style strength control as a regression learning between style control parameter $\alpha$ and style strength of output image. In this aspect, the feed-forward networks from previous methods are strongly biased to full style strength ($\alpha = 1.0$), lacking training data for intermediate style strength ($\alpha < 1.0$). Therefore, here, we alternatively propose an unbiased learning of style transfer network by using additional training data and style loss for $\alpha < 1.0$ in training phase. As shown in fig.\ref{fig:concept}(b), we use the unbiased training data and corresponding loss for $\alpha = 0.0$ to make the trained network to reconstruct input content image when $\alpha = 0.0$ as well as to generate target stylized image when $\alpha = 1.0$. This unbiased training also helps in selecting an appropriate weight for style loss by reducing network's sensitivity to the weight of style loss. Moreover, with additional anchor data and the corresponding loss for $0.0 < \alpha < 1.0$ as shown in fig.\ref{fig:concept}(c), our method allows the network to learn the desired regression between style control parameter and output style strength consistent with a specific style interpolation function in encoded feature space. Figure \ref{fig:network architecture} shows the whole network architecture and losses of our method.

In the remained of this paper, details of our unbiased style regression learning method will be described in sec.\ref{sec:Method}, experimental results and analysis of verifying our method will be presented in sec.\ref{sec:Experiments}, and we will conclude this work in sec.\ref{sec:Conclusion}.

\subsection{Related works}
\label{sec:related works}
For the first neural approach for image style transfer, Gatys \etal~\cite{Gatys_2016_CVPR} adopted a part of VGG-net~\cite{Simonyan14c}, which is a pre-trained convolutional neural network for image classification task, as a feature extractor for content and style of an image. They generated output image similar to an input content image in content and to a target style image in style by updating pixel values to minimize the summation of content difference and style difference using an online gradient-based optimization technique. For the content and style similarity measure, they used the difference in VGG feature space as the content loss and the difference in Gram space of the feature as the style loss. Their method resulted in a good quality of transferred style but it had a very slow image generating speed because of its online optimization scheme.

By inserting a feed-forward network between the input and output images~\cite{Johnson_2016_ECCV,Ulyanov_2016_ICML,Ulyanov_2017_CVPR}, the problem of a very slow image generating speed was solved. The feed-forward network was trained to generate output stylized image from input content image for a target style. This changed the online pixel optimization process \cite{Gatys_2016_CVPR} into an offline network training and sped up the image generating process as a network feed-forward calculation.

Soon after, implementation of modified instance normalization layer~\cite{Dumoulin_2017_ICLR,Ghiasi_2017_BMVC,Huang_2017_ICCV,Li_2017_NIPS} allowed the trained network to embed multiple or arbitrary styles and to generate the output image of mixed style or intermediate style strength. Dumoulin \etal~\cite{Dumoulin_2017_ICLR} used learnable affine parameters for multiple styles in their conditional instance normalization (CIN) layer to efficiently switch the style of the output image to a desired style by changing 2nd order statistics in VGG feature space. In addition, they proposed a simple style interpolation technique to generate an output image of mixed style by linearly interpolating affine parameters of embedded styles. Huang and Belongie~\cite{Huang_2017_ICCV} proposed the alternative adaptive instance normalization (AdaIN) layer for transferring the style of an unseen target image to the content image. Instead of using learnable parameters, they used human-designed parameters, mean and standard deviation, of VGG feature for changing feature statistics. Their method also used linear interpolation of the mean and standard deviation in their AdaIN layer to control style strength of the output image.

Li \etal~\cite{Li_2017_NIPS} used a correlation-aware feature alignment technique called whitening and coloring (WCT), also known as correlation alignment (CORAL)~\cite{Sun_2016_AAAI,Sun_2016_ECCV} in object classification. Here, they used covariance instead of standard deviation to consider inter-channel correlation in feature statistics. They also used linear interpolation technique for style strength control as previously demonstrated \cite{Huang_2017_ICCV} but achieved a better quality of the reconstructed content image with zero style strength because they trained decoder network in the manner of minimizing both pixel reconstruction loss and feature loss. However, they did not deal with exact regression between style control parameter and style strength of output image.

Additionally, generative adversarial network (GAN) approaches similar to Pix2pix~\cite{Isola_2017_CVPR}, CycleGAN~\cite{Zhu_2017_ICCV}, and BicycleGAN~\cite{Zhu_2017_NIPS}, also dealt with image style transfer as an application of their image-to-image translation task. These methods relieved the requirement of well-defined loss for image style difference and training image pairs which are necessary for encoder/decoder network. While these approaches achieved a high quality of the generated image by focusing on realistic image generation, these did not focus on style strength control.

\section{Method}
\label{sec:Method}
Our unbiased image style transferring method consists of two strategies. One is unbiased network learning to generate the zero-style image of content image style. The other is regression specification to control style strength of output image in desired characteristic function between style control parameter and output style strength.

\subsection{Unbiased learning for unbiased and stable style transfer}
\label{sec:unbiased learning for stable style transfer}
\begin{figure}
\includegraphics[trim={0 12cm 0 0},width=1.0\textwidth]{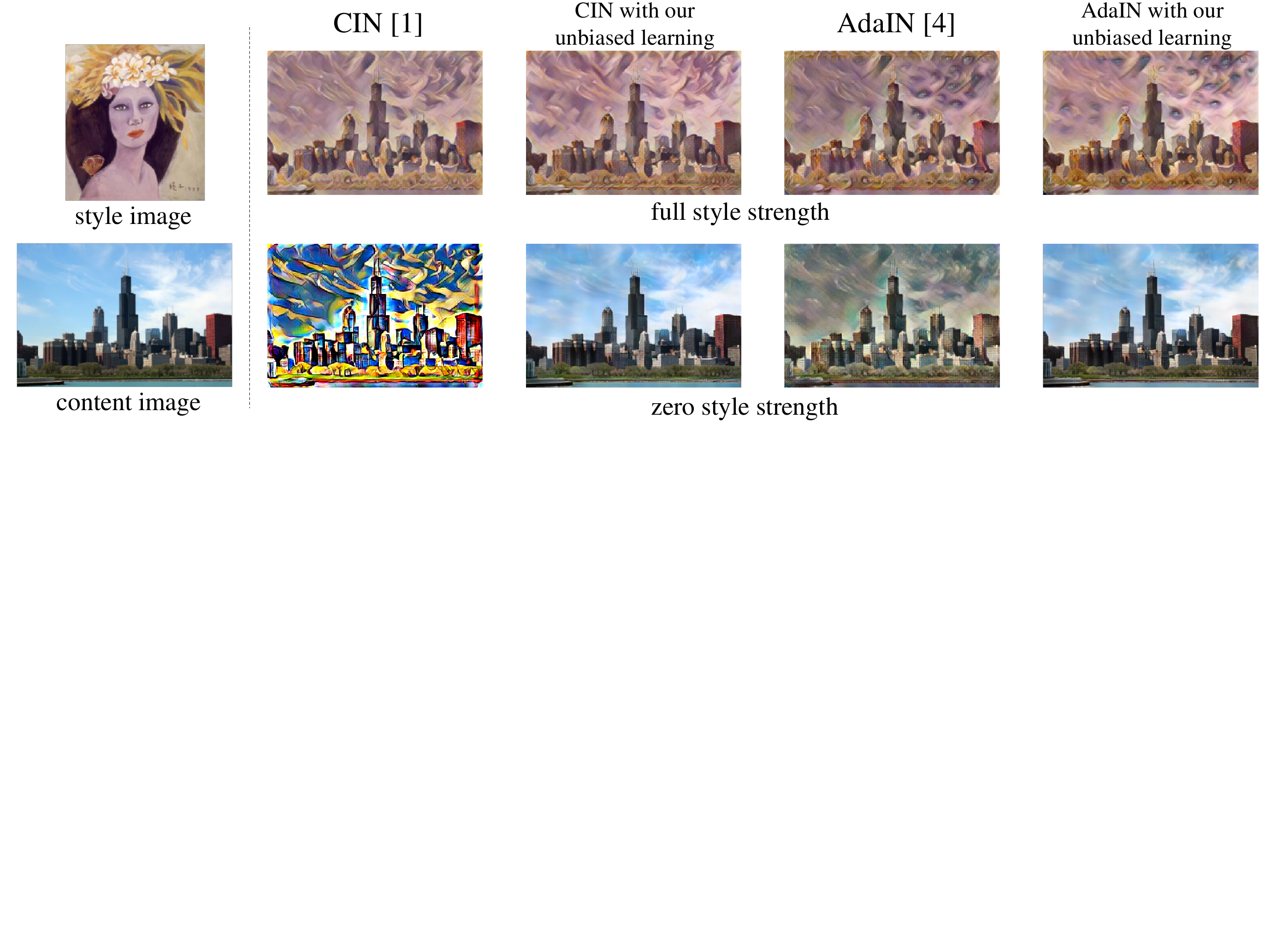}
\caption{Unbiased learning vs. biased learning with a small set (22 images) of target styles: The generated full style images of all methods have the style very similar to the style image as shown in the first row of images. However, the generated zero-style images of CIN and AdaIN (the second and the fourth images on the bottom row) are much different from the content image while the images of our unbiased methods (the third and the last images on the bottom row) have the style similar to the content image.}
\label{fig:biased vs unbiased}
\end{figure}

Output images of the previous feed-forward networks~\cite{Dumoulin_2017_ICLR,Huang_2017_ICCV} with the style control parameter $\alpha = 0.0$ are not same to the content image but an image of some biased style as shown in fig.\ref{fig:biased vs unbiased}. The biased images of zero style strength occurred because the decoder networks of the previous methods were trained only with the biased data of \{content image, target style image\} pairs which are corresponding to full style strength ($\alpha = 1.0$). As the result, the biased decoder does not guarantee the unbiased output of zero style strength when style control parameter is zero.

As a simple but effective way to solve this problem of the biased decoder, we add the unbiased data of \{content image, content image\} pairs in every batch iteration of network training phase as shown in fig.\ref{fig:concept}(b) and the corresponding unbiased loss $L_{unbiased}$ as fig.\ref{fig:network architecture}. This means that the losses corresponding to the unbiased data, i.e., unbiased content loss ${L_{u}}_{content}$, unbiased style loss ${L_{u}}_{style}$, and unbiased total variation loss ${L_{u}}_{tv}$, are added to the biased loss $L_{biased}$ which was calculated as a weighted summation of the losses of biased data, i.e., $L_{content}$ \cite{Gatys_2016_CVPR,Dumoulin_2017_ICLR}, $L_{style}$ \cite{Huang_2017_ICCV}, and $L_{tv}$ \cite{Gatys_2016_CVPR}. These unbiased losses give additional constraint to the decoder network to generate an output image of original content style while the biased losses encourage the decoder network to be optimized to generate an output image of target style. Total loss considering our unbiased loss is represented as eq.\ref{eq:our loss}.
\begin{multline}
L_{total} = L_{biased} + L_{unbiased}, \\
L_{biased} = w_c \cdot L_{content}(I,\:I_c) + w_s \cdot L_{style}(I,\:I_s) + w_{t} \cdot L_{tv}(I), \\
L_{unbiased} = w_c \cdot {L_{u}}_{content} + w_s \cdot {L_{u}}_{style} + w_{t} \cdot {L_{u}}_{tv} + w_r \cdot L_{reconstruct}, \\
{L_{u}}_{content} = L_{content}(I_u,\:I_c),\:{L_{u}}_{style} = L_{style}(I_u,\:I_s),\:{L_{u}}_{tv} = L_{tv}(I_u),\:L_{reconstruct} = || I_u - I_c ||_1.
\label{eq:our loss}
\end{multline}
where we add additional $L1$ loss, $L_{reconstruct}$ (eq.\ref{eq:our loss}), between the zero-style image $I_u$ and the content image $I_c$ into the total loss $L_{unbiased}$ to reconstruct content image when style strength is zero. $L_{reconstruct}$ is consistent to that of \cite{Li_2017_NIPS} where $L2$ loss was used to train the decoder network. However, using $L2$ loss is known for blurred reconstructed image~\cite{DBLP:journals/corr/MathieuCL15,Pathak_2016_CVPR,Zhang_2016_ECCV,Isola_2017_CVPR}, and \cite{Li_2017_NIPS} used only unbiased losses of unbiased data (content images) to train the decoder network. In the loss equation form, \cite{Li_2017_NIPS} is a specific case of our unbiased learning scheme because eliminating the biased losses and style losses reduces our total loss (eq.\ref{eq:our loss}) into that of \cite{Li_2017_NIPS}.

\subsection{Regression specification for style control}
\label{sec:regression specification for style control}
As shown in fig.\ref{fig:concept}(a), training a network only with the biased data of \{content image, target style image\} pairs cannot guarantee to learn a linear regression between style control parameter and style strength of output image which was used for style interpolation in the previous image style transferring methods~\cite{Dumoulin_2017_ICLR,Huang_2017_ICCV,Li_2017_NIPS}.

To learn a specific regression between style control parameter and style strength of output image, we need to use additional anchor data as shown as green dots in fig.\ref{fig:concept}(c) and corresponding anchor losses $L_{anchor}$ as shown in fig.\ref{fig:network architecture} for intermediate values of style control parameter $\alpha$. The anchor loss $L_{anchor}$ is represented in eq.\ref{eq:anchor loss} in the same manner of $L_{biased}$ in eq.\ref{eq:our loss}. Here, the anchor-style loss ${L_{a}}_{style}$ is the style distance between the output anchor image $I_{\alpha}$ and target anchor-style image $I_s(\alpha)$. However, it is not possible to calculate ${L_{a}}_{style}$ directly from the images because we do not have the target anchor-style image $I_s(\alpha)$. Therefore, as an alternative of $I_s(\alpha)$, we use the linear interpolated style feature of full style feature $f_s(I_s)$ and zero-style feature $f_s(I_c)$ as the target anchor-style feature. Then, the anchor-style loss can be calculated as the $L2$ distance between the target anchor-style feature and the output anchor-style feature $f_s(I_{\alpha})$ as in eq.\ref{eq:anchor loss}.
\begin{multline}
L_{anchor}(\alpha) = w_c \cdot {L_a}_{content} + w_s \cdot {L_a}_{style} + w_{t} \cdot {L_a}_{tv}, \\
{L_a}_{content} = L_{content}(I_{\alpha},\:I_c),\:{L_a}_{tv} = L_{tv}(I_{\alpha}), \\
{L_a}_{style} = L_{style}(I_{\alpha},\:I_s(\alpha)) = || f_s(I_{\alpha}) - (\alpha \cdot f_s(I_s) + (1-\alpha) \cdot f_s(I_c)) ||_2.
\label{eq:anchor loss}
\end{multline}
This anchor loss for desired value of $\alpha$ is added to the total loss of eq.\ref{eq:our loss} in every iteration of the training phase. Once a network is trained as a linear regressor, then we can specify arbitrary regression by using a desired characteristic function $f(\alpha)$ instead of $\alpha$ in transformer of the network (fig.\ref{fig:network architecture}).

\section{Experiments}
\label{sec:Experiments}
In this section, we will analyze our unbiased learning and regression specifying methods experimentally in the aspect of loss and image quality. And we will prove the benefits of our method by comparing to the previous image style transferring methods.

\begin{figure}
\includegraphics[trim={0 13.5cm 0 0},width=1.0\textwidth]{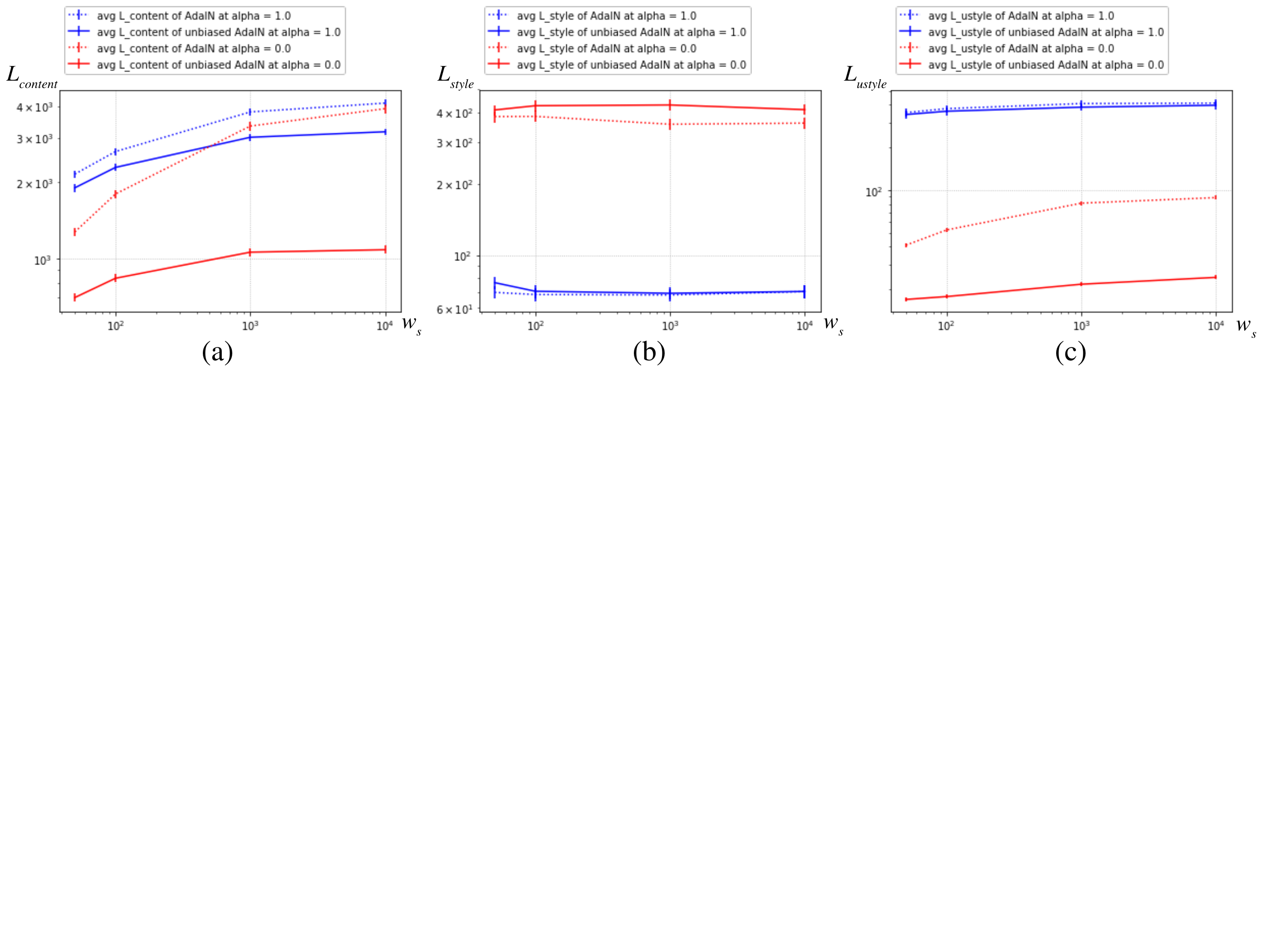}
\caption{Loss vs. style weight: The represented losses were averaged over 100 style transfer tests and drawn with standard deviation times 0.1. (a) AdaIN with our unbiased learning scheme (solid lines) has a much lower content loss (indicating better content preserving property) than the original AdaIN method (dashed lines) for a wide range of style loss weight.; (b) Style loss (distance from the target style) of our unbiased method is almost same as that of AdaIN method. This indicates the similar ability to maintain style.; (c) Average unbiased style loss (distance from the content style) of our unbiased method is much lower at $\alpha=0.0$ than that of AdaIN method, indicating a better content style reconstruction.}
\label{fig:style weight vs losses}
\end{figure}

\subsection{Experimental Setup}
\label{sec:experimental setup}
We used the encoder-transformer-decoder architecture of AdaIN~\cite{Huang_2017_ICCV} as the common network configuration but VGG16 feature extractor as the encoder and its mirrored network as the decoder respectively as shown in fig.\ref{fig:network architecture}. The output tensors of \{relu1\_2, relu2\_2, relu3\_3, relu4\_3\} layers were used as the style features and that of \{relu3\_3\} layer as the content feature. This follows the layer configuration of \cite{Johnson_2016_ECCV}, which uses VGG16 feature extractor in loss calculation. We set the weights of losses as $w_c=1.0$, $w_t=10^{-3}$, $w_r=10^2 \cdot w_s$, and varying weights ($w_s=50, 10^2, 10^3, 10^4$) to analyze how the learned networks work as the weight of style loss increases.

For training data, we used MS COCO train2014 dataset~\cite{Lin_2014_ECCV} as content images and the training dataset of painter by numbers~\cite{painter_by_numbers} as a large set of target style images. Additionally, we used our collection of 22 style images as a small set of target style images to analyze network performance as the number of embedded style increases and to compare our method to CIN~\cite{Dumoulin_2017_ICLR} which can be applied to a small number of target styles. Those images were resized into 256 pixels in short side and cropped into 240 by 240 pixels for data augmentation while containing a reasonable amount of image content. With those training images, the networks of CIN layer or AdaIN layer with or without our unbiased learning scheme were trained by Adam optimizer~\cite{Kingma_2015_ICLR} with learning rate $10^{-4}$ (with smaller learning rate $10^{-6}$ when $w_s=10^4$), batch size 4 and epoch number 4 on Pytorch v0.3.1 framework with CUDA v9.0, CuDNN v7.0, and NVIDIA TITAN-X Pascal. In test phase, we used MS COCO test2014 dataset~\cite{Lin_2014_ECCV} and test dataset of painter by numbers~\cite{painter_by_numbers} as the content images and the target style images respectively, and all the test images were resized into 256 pixels in short side without cropping before fed into the networks.

\subsection{Results of unbiased learning}
\label{sec:results of unbiased learning}
As shown in fig.\ref{fig:biased vs unbiased}, the networks with CIN layer or AdaIN layer trained by using the previous biased training schemes~\cite{Dumoulin_2017_ICLR,Huang_2017_ICCV} with a small set of style images generated full style images of high style quality but heavily biased zero-style images. In contrast, the trained networks with our unbiased scheme generated unbiased zero-style images while maintaining almost the same quality of full style images.

\begin{figure}
\includegraphics[trim={0 3cm 0 0},width=1.0\textwidth]{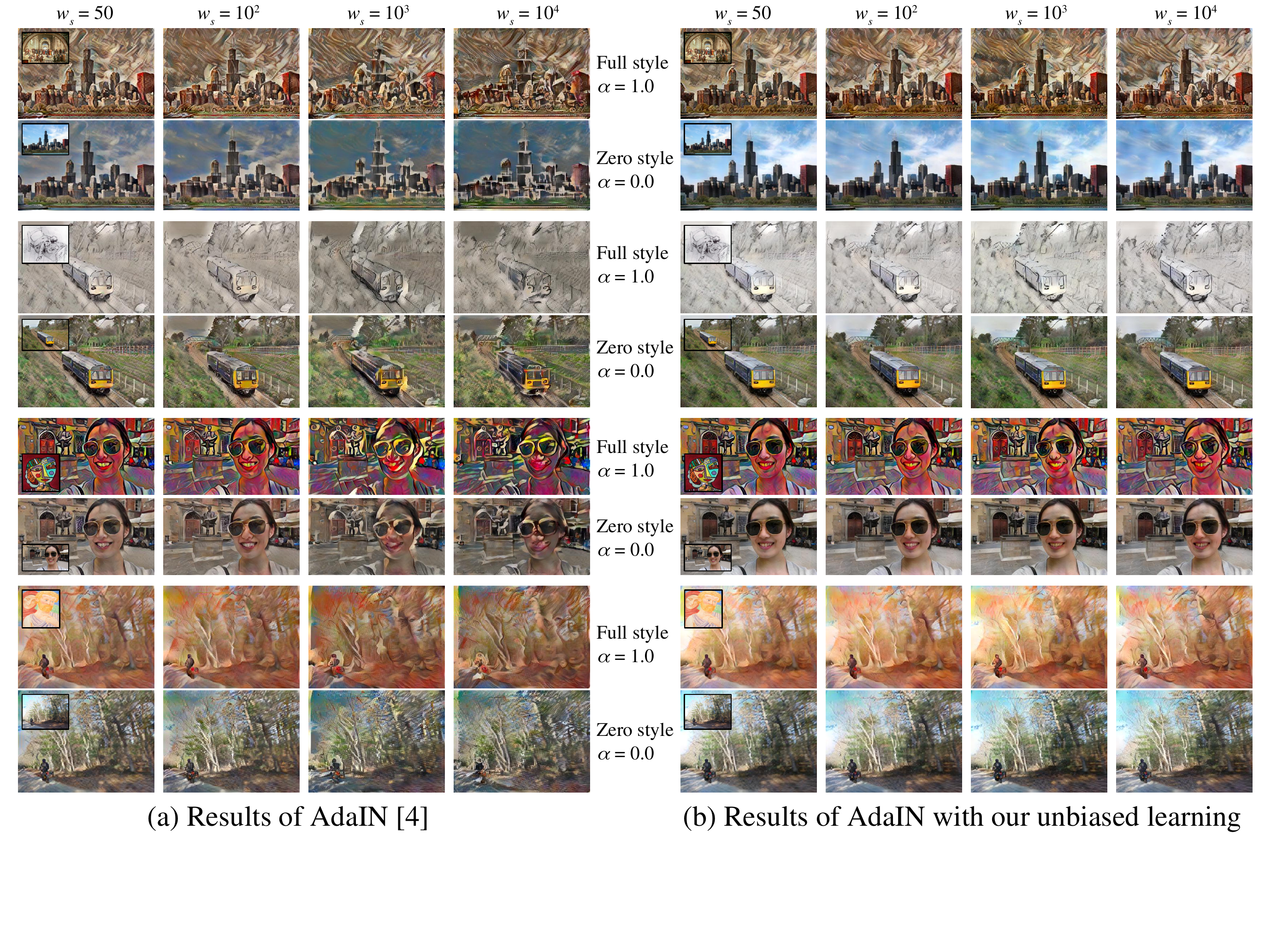}
\caption{Comparison of AdaIN with or without our unbiased learning scheme for varying style weight: (a) AdaIN generated full style images of the target style image for small style loss weight $w_s=50$ (left-most images on odd rows). However, for high style loss weight $w_s \geq 10^2$, the full style images are degraded in both content and style (images on odd rows). The zero-style images (images on even rows) are much different from the content images.; (b) AdaIN with our unbiased learning scheme generated full style images of target style and zero-style images very similar to the content image for a wide range of style loss weight $50 \leq w_s \leq 10^4$. The test images were not seen during the training phase.}
\label{fig:unbiased vs biased AdaIN}
\end{figure}

For more generalized performance comparison, we trained several networks with a large set of style images and with varying weights of style loss. Afterword, we measured the average values of content losses $L_{content}$, style losses $L_{style}$, and unbiased style losses $L_{ustyle}$ in eq.\ref{eq:our loss} for test style transfer with 100 pairs of \{unseen content image, unseen target style image\}. Figure.\ref{fig:style weight vs losses} shows the measured average losses and its standard deviations times 0.1. When $\alpha=1.0$ (full style transfer), our unbiased learning scheme achieved the smaller average content loss than that of the original AdaIN (blue lines on fig.\ref{fig:style weight vs losses}(a)) while maintaining almost the same average style loss and unbiased style loss of original AdaIN (blue lines on fig.\ref{fig:style weight vs losses}(b), (c)). This means that the fully stylized images of our method (odd rows of fig.\ref{fig:unbiased vs biased AdaIN}(b)) have less degradation in content than those of the previous method (odd rows of fig.\ref{fig:unbiased vs biased AdaIN}(a)) for the same stylization quality. When $\alpha=0.0$ (zero-style transfer), our unbiased learning scheme achieved a much smaller average content loss and unbiased style loss than those of the original AdaIN (red lines on fig.\ref{fig:style weight vs losses}(a), (c)) while maintaining the higher average style loss (red lines on fig.\ref{fig:style weight vs losses}(b)). This means that the zero stylized images of our method (even rows of fig.\ref{fig:unbiased vs biased AdaIN}(b)) are almost reconstructed into the original content images while those of the previous method (even rows of fig.\ref{fig:unbiased vs biased AdaIN}(a)) are quite different from the original content images.

As the weight of style loss $w_s$ increases, content losses at $\alpha=1.0,\:0.0$ and unbiased style loss at $\alpha=0.0$ also increase as shown in fig.\ref{fig:style weight vs losses}(a) and (c). However, the increment is much smaller with our unbiased learning scheme compared to that of the original AdaIN. This means that our method achieved stable stylization performance by maintaining desired content and style of output image insensitive to the large style weight. This stableness in stylization is also verified in fig.\ref{fig:unbiased vs biased AdaIN}. The full style and zero style results of original AdaIN (fig.\ref{fig:unbiased vs biased AdaIN}(a)) shows good output style quality with small style weights but degraded style quality with large style weights. In contrast, the results of our unbiased scheme (fig.\ref{fig:unbiased vs biased AdaIN}(b)) shows a comparable quality of full style and zero-style images stable with a large range of style weight variation.

\subsection{Results of specifying style regression}
\label{sec:results of specifying style regression}
To verify how our regression specification method works, we trained additional style transfer networks with AdaIN layer for linear style regression learning by using two additional anchor losses (eq.\ref{eq:anchor loss}) at $\alpha=\frac{1}{3},\:\frac{2}{3}$ for intermediate target styles to the total loss of eq.\ref{eq:our loss} (we could add only two additional anchors because of memory limitation on GPU device). Then, we calculated the average content losses $L_{content}$ and the average anchor-style losses ${L_{a}}_{style}$ of the output stylized images of the varying style control parameter $\alpha$ by feeding 100 test pairs of \{content image, target style image\} into the trained networks.

\begin{figure}
\includegraphics[trim={0 14.5cm 0 0},width=1.1\textwidth]{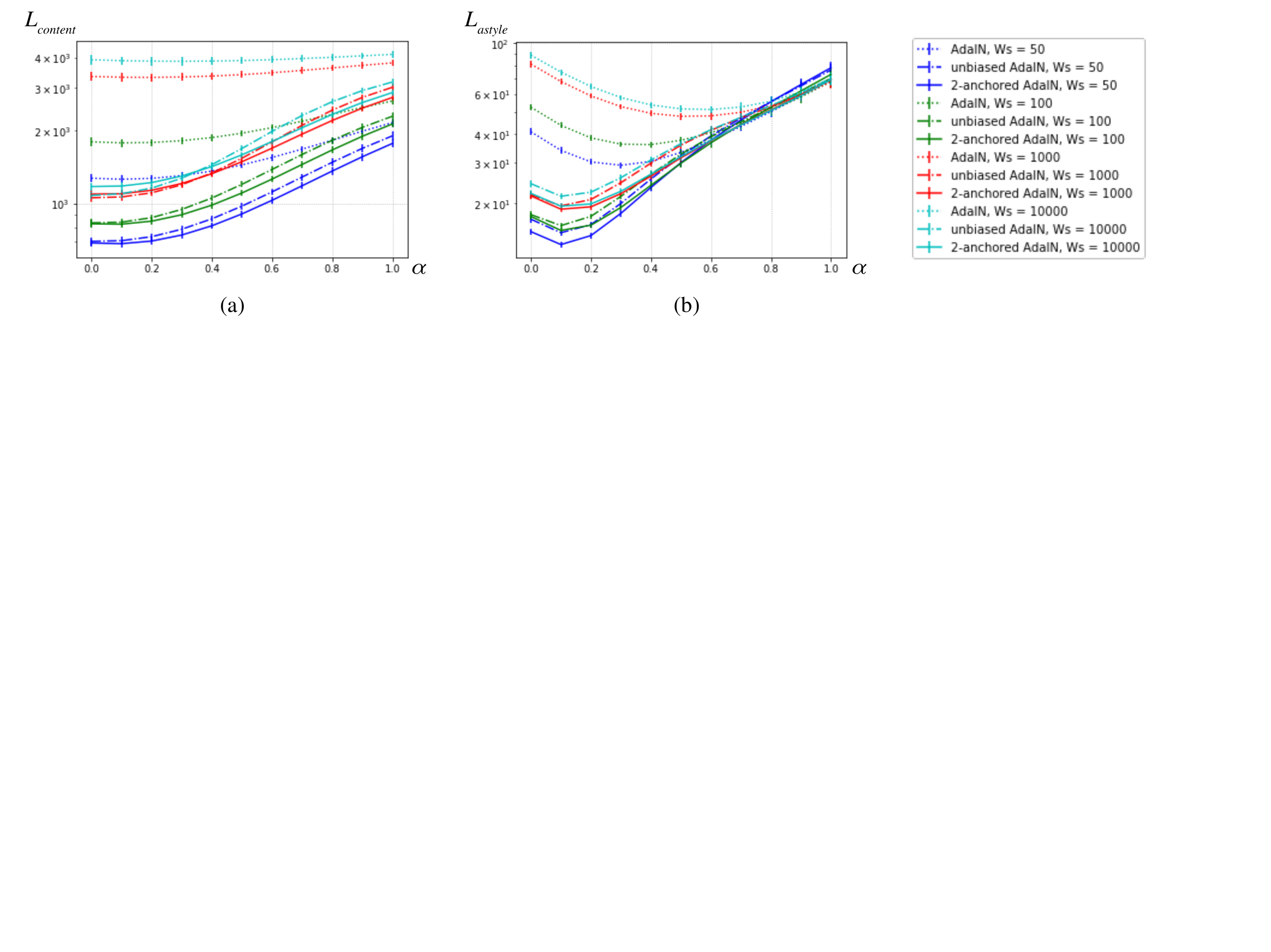}
\caption{Loss vs. style control parameter: The above losses were averaged over 100 style transfer tests and drawn with standard deviation times 0.1. (a) AdaIN with our unbiased learning scheme (dash-dot lines) and with additional two anchor data (solid lines) have smooth transition in content loss for varying style control parameter $\alpha$ while maintaining the much lower values than the original AdaIN method (dashed lines).; (b) The anchor-style loss ${L_a}_{style}$ (distance from the target anchor-style of eq.\ref{eq:anchor loss}) of our unbiased method with additional two anchor data (solid lines) has lower value (indicating better regression matching) than that without anchor data (dash-dot lines) and much lower value than that of AdaIN without our unbiased learning scheme (dashed lines).}
\label{fig:style control parameter vs losses}
\end{figure}

Figure~\ref{fig:style control parameter vs losses} shows how the losses change as the style control parameter changes. As shown in fig.\ref{fig:style control parameter vs losses}(a), The content losses of our unbiased method (dash-dot lines) and 2-anchored method (solid lines) smoothly increases as the style control parameter increases, maintaining much lower value than that of original AdaIN (dashed lines). This is the expected content-style trade-off in style strength control but shows that considering additional anchor data of intermediate style did not degrade the stable content-preserving property of our unbiased learning scheme and even achieved more stable (slightly lower content loss) than the unbiased method did as $\alpha$ increases.

As shown in fig.\ref{fig:style control parameter vs losses}(b), anchor-style loss of our 2-anchored learning method (solid lines) is lower than that of our unbiased learning method (dash-dot lines) for almost all $\alpha$ values and much lower than that of original AdaIN (dashed lines) for lower $\alpha$ values. This means that the learned style regression with additional anchor losses is closer to the desired linear regression than those of the original AdaIN and the unbiased AdaIN.

\begin{figure}
\includegraphics[trim={0 5cm 0 0},width=1.0\textwidth]{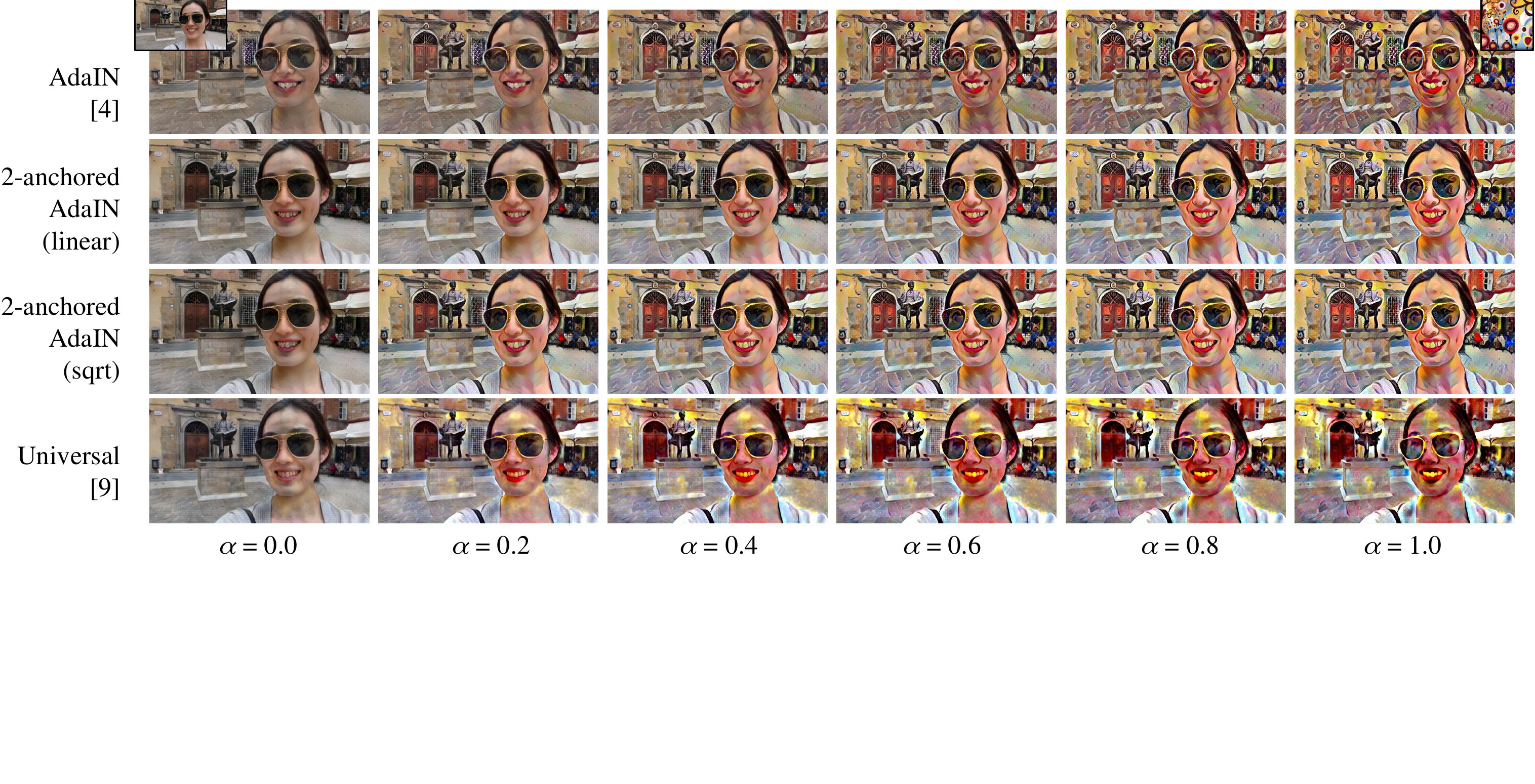}
\caption{Style strength control when style loss weight $w_s=50$: AdaIN~\cite{Huang_2017_ICCV} generated the biased images for all $\alpha$ values (first row). Our unbiased learning scheme with two additional anchor data shows smooth linear (second row) or square-root (third row) style transitions from the well-reconstructed content images of $\alpha=0.0$ to the full style images of $\alpha=1.0$. Universal~\cite{Li_2017_NIPS} (bottom row) shows saturated images for $\alpha > 0.6$ and blurred images.}
\label{fig:unbiased vs anchor AdaIN}
\end{figure}

Figure~\ref{fig:unbiased vs anchor AdaIN} shows the intermediate stylized images of the previous and our methods. The results of our unbiased scheme with 2 anchor data are presenting the unbiased smooth style transitions for regressions of $f(\alpha)=\alpha$ (second row of fig.\ref{fig:unbiased vs anchor AdaIN}) and $f(\alpha)=\sqrt{\alpha}$ (third row of fig.\ref{fig:unbiased vs anchor AdaIN}) as the style control parameter $\alpha$ changes while the original AdaIN is presenting the style transition starting from the biased zero-style image (first row of fig.\ref{fig:unbiased vs anchor AdaIN}). The result of Universal~\cite{Li_2017_NIPS} (bottom row of fig.\ref{fig:unbiased vs anchor AdaIN}) seems to present good style transition especially in pixel color but shows blurred images caused by its $L2$ reconstruction loss, lack of stroke patterns, and style saturation for $\alpha > 0.6$ caused by using only content images in detector learning. Here, the style transfer networks of Universal were trained as a cascade of 3 networks under the common network configuration described in sec.\ref{sec:experimental setup}.

\section{Conclusion}
\label{sec:Conclusion}
In this paper, we proposed the unbiased learning scheme and the style regression specifying technique for fast image style transfer based on feed-forward neural networks. Our unbiased learning scheme used biased loss and unbiased loss simultaneously in network training and achieved a stable style transfer with a wide range of style loss weight and a content-preserving style transfer enough to reconstruct content image when style control parameter is zero. Moreover, by considering additional anchor data of intermediate styles, our method improved to learn a regression between output style strength and style control parameter closer to the linear regression. This resulted in arbitrary regression of style strength by using a desired regression function of style control parameter in the transformer of style transfer networks. These achievements were verified experimentally by analyzing the losses and the generated images from a number of trained networks for the state-of-the-art methods and ours with a wide range of style loss weights and varying style control parameters.

\bibliography{egbib}

\begin{thebibliography}{21}
\providecommand{\natexlab}[1]{#1}
\providecommand{\url}[1]{\texttt{#1}}
\expandafter\ifx\csname urlstyle\endcsname\relax
  \providecommand{\doi}[1]{doi: #1}\else
  \providecommand{\doi}{doi: \begingroup \urlstyle{rm}\Url}\fi

\bibitem[Dumoulin et~al.(2017)Dumoulin, Shlens, and Kudlur]{Dumoulin_2017_ICLR}
Vincent Dumoulin, Jonathon Shlens, and Manjunath Kudlur.
\newblock A learned representation for artistic style.
\newblock In \emph{International Conference on Learning Representations
  (ICLR)}, Apr 2017.

\bibitem[Gatys et~al.(2016)Gatys, Ecker, and Bethge]{Gatys_2016_CVPR}
Leon~A. Gatys, Alexander~S. Ecker, and Matthias Bethge.
\newblock Image style transfer using convolutional neural networks.
\newblock In \emph{The IEEE Conference on Computer Vision and Pattern
  Recognition (CVPR)}, June 2016.

\bibitem[Ghiasi et~al.(2017)Ghiasi, Lee, Kudlur, Dumoulin, and
  Shlens]{Ghiasi_2017_BMVC}
Golnaz Ghiasi, Honglak Lee, Manjunath Kudlur, Vincent Dumoulin, and Jonathon
  Shlens.
\newblock Exploring the structure of a real-time, arbitrary neural artistic
  stylization network.
\newblock In \emph{British Machine Vision Conference (BMVC)}, Sep 2017.

\bibitem[Huang and Belongie(2017)]{Huang_2017_ICCV}
Xun Huang and Serge Belongie.
\newblock Arbitrary style transfer in real-time with adaptive instance
  normalization.
\newblock In \emph{The IEEE International Conference on Computer Vision
  (ICCV)}, Oct 2017.

\bibitem[Isola et~al.(2017)Isola, Zhu, Zhou, and Efros]{Isola_2017_CVPR}
Phillip Isola, Jun-Yan Zhu, Tinghui Zhou, and Alexei~A. Efros.
\newblock Image-to-image translation with conditional adversarial networks.
\newblock In \emph{The IEEE Conference on Computer Vision and Pattern
  Recognition (CVPR)}, July 2017.

\bibitem[Johnson et~al.(2016)Johnson, Alahi, and Fei-Fei]{Johnson_2016_ECCV}
Justin Johnson, Alexandre Alahi, and Li~Fei-Fei.
\newblock Perceptual losses for real-time style transfer and super-resolution.
\newblock In \emph{European Conference on Computer Vision (ECCV)}, Oct 2016.

\bibitem[Kingma and Ba(2015)]{Kingma_2015_ICLR}
Diederik~P. Kingma and Jimmy Ba.
\newblock Adam: {A} method for stochastic optimization.
\newblock In \emph{International Conference on Learning Representations
  (ICLR)}, May 2015.

\bibitem[Li et~al.(2017{\natexlab{a}})Li, Wang, Liu, and Hou]{Li_2017_IJCAI}
Yanghao Li, Naiyan Wang, Jiaying Liu, and Xiaodi Hou.
\newblock Demystifying neural style transfer.
\newblock In \emph{IJCAI}, 2017{\natexlab{a}}.

\bibitem[Li et~al.(2017{\natexlab{b}})Li, Fang, Yang, Wang, Lu, and
  Yang]{Li_2017_NIPS}
Yijun Li, Chen Fang, Jimei Yang, Zhaowen Wang, Xin Lu, and Ming-Hsuan Yang.
\newblock Universal style transfer via feature transforms.
\newblock In \emph{Advances in Neural Information Processing Systems (NIPS)},
  Dec 2017{\natexlab{b}}.

\bibitem[Lin et~al.(2014)Lin, Maire, Belongie, Hays, Perona, Ramanan, Dollár,
  and Zitnick]{Lin_2014_ECCV}
Tsung-Yi Lin, Michael Maire, Serge Belongie, James Hays, Pietro Perona, Deva
  Ramanan, Piotr Dollár, and C.~Lawrence Zitnick.
\newblock Microsoft {COCO:} common objects in context.
\newblock In \emph{European Conference on Computer Vision (ECCV)}, Sep 2014.

\bibitem[Mathieu et~al.(2015)Mathieu, Couprie, and
  LeCun]{DBLP:journals/corr/MathieuCL15}
Micha{\"{e}}l Mathieu, Camille Couprie, and Yann LeCun.
\newblock Deep multi-scale video prediction beyond mean square error.
\newblock \emph{CoRR}, abs/1511.05440, 2015.
\newblock URL \url{http://arxiv.org/abs/1511.05440}.

\bibitem[Nichol(2016)]{painter_by_numbers}
Kiri Nichol.
\newblock Kaggle dataset: Painter by numbers, 2016.
\newblock \url{https://www.kaggle.com/c/painter-by-numbers}.

\bibitem[Pathak et~al.(2016)Pathak, Kr\"ahenb\"uhl, Donahue, Darrell, and
  Efros]{Pathak_2016_CVPR}
Deepak Pathak, Philipp Kr\"ahenb\"uhl, Jeff Donahue, Trevor Darrell, and Alexei
  Efros.
\newblock Context encoders: Feature learning by inpainting.
\newblock In \emph{The IEEE Conference on Computer Vision and Pattern
  Recognition (CVPR)}, June 2016.

\bibitem[Simonyan and Zisserman(2014)]{Simonyan14c}
K.~Simonyan and A.~Zisserman.
\newblock Very deep convolutional networks for large-scale image recognition.
\newblock \emph{CoRR}, abs/1409.1556, 2014.

\bibitem[Sun and Saenko(2016)]{Sun_2016_ECCV}
Baochen Sun and Kate Saenko.
\newblock Deep coral: Correlation alignment for deep domain adaptation.
\newblock In \emph{ECCV 2016 Workshops}, 2016.

\bibitem[Sun et~al.(2016)Sun, Feng, and Saenko]{Sun_2016_AAAI}
Baochen Sun, Jiashi Feng, and Kate Saenko.
\newblock Return of frustratingly easy domain adaptation.
\newblock In \emph{Proceedings of the Thirtieth AAAI Conference on Artificial
  Intelligence}, AAAI'16, pages 2058--2065. AAAI Press, 2016.
\newblock URL \url{http://dl.acm.org/citation.cfm?id=3016100.3016186}.

\bibitem[Ulyanov et~al.(2016)Ulyanov, Lebedev, Vedaldi, and
  Lempitsky]{Ulyanov_2016_ICML}
Dmitry Ulyanov, Vadim Lebedev, Andrea Vedaldi, and Victor~S. Lempitsky.
\newblock Texture networks: Feed-forward synthesis of textures and stylized
  images.
\newblock In \emph{International Conference on Machine Learning (ICML)}, June
  2016.

\bibitem[Ulyanov et~al.(2017)Ulyanov, Vedaldi, and
  Lempitsky]{Ulyanov_2017_CVPR}
Dmitry Ulyanov, Andrea Vedaldi, and Victor Lempitsky.
\newblock Improved texture networks: Maximizing quality and diversity in
  feed-forward stylization and texture synthesis.
\newblock In \emph{The IEEE Conference on Computer Vision and Pattern
  Recognition (CVPR)}, July 2017.

\bibitem[Zhang et~al.(2016)Zhang, Isola, and Efros]{Zhang_2016_ECCV}
Richard Zhang, Phillip Isola, and Alexei~A Efros.
\newblock Colorful image colorization.
\newblock In \emph{European Conference on Computer Vision (ECCV)}, Oct 2016.

\bibitem[Zhu et~al.(2017{\natexlab{a}})Zhu, Park, Isola, and
  Efros]{Zhu_2017_ICCV}
Jun-Yan Zhu, Taesung Park, Phillip Isola, and Alexei~A. Efros.
\newblock Unpaired image-to-image translation using cycle-consistent
  adversarial networks.
\newblock In \emph{The IEEE International Conference on Computer Vision
  (ICCV)}, Oct 2017{\natexlab{a}}.

\bibitem[Zhu et~al.(2017{\natexlab{b}})Zhu, Zhang, Pathak, Darrell, Efros,
  Wang, and Shechtman]{Zhu_2017_NIPS}
Jun-Yan Zhu, Richard Zhang, Deepak Pathak, Trevor Darrell, Alexei~A Efros,
  Oliver Wang, and Eli Shechtman.
\newblock Toward multimodal image-to-image translation.
\newblock In \emph{Advances in Neural Information Processing Systems (NIPS)},
  Dec 2017{\natexlab{b}}.

\end{thebibliography}
\end{document}